# CloudifierNet - Deep Vision Models for Artificial Image Processing

Andrei Ionut Damian[a], Laurentiu Piciu[a], Alexandru Purdila[a], Nicolae Tapus[b]

*[a]Lummetry.AI, Bucharest, Romania; [b] University "Politehnica" Bucharest, Romania*

**Abstract**

Today, more and more, it is necessary that most applications and documents developed in previous or current technologies to be accessible online on cloud-based infrastructures. That's why the migration of legacy systems including their hosts of documents to new technologies and online infrastructures, using modern Artificial Intelligence techniques, is absolutely necessary.

With the advancement of Artificial Intelligence and Deep Learning with its multitude of applications, a new area of research is emerging – that of automated systems development and maintenance. The underlying work objective that led to this paper aims to research and develop truly intelligent systems able to analyze user interfaces from various sources and generate real and usable inferences ranging from architecture analysis to actual code generation. One key element of such systems is that of artificial scene detection and analysis based on deep learning computer vision systems. Computer vision models and particularly deep directed acyclic graphs based on convolutional modules are generally constructed and trained based on natural images datasets. Due to this fact, the models will develop during the training process natural image feature detectors apart from the base graph modules that will learn basic primitive features. In the current paper, we will present the base principles of a deep neural pipeline for computer vision applied to artificial scenes (scenes generated by user interfaces or similar). Finally, we will present the conclusions based on experimental development and benchmarking against state-of-the-art transfer-learning implemented deep vision models

## 1. Introduction

In our days, the companies need to offer collaborative online experience. Today's most competitive companies have professionals from different fields who are working collaborative and contribute, sharing experiences, participating in decision making, and influencing changes inside or outside of the company [1]. Subsequently, the quality of the decision depends on the available data imposing the need to use all possibly useful data no matter which is the initial format, stocking or source.

Within this area of application of Artificial Intelligence, a broad range of computer users – from IT maintenance personnel to software developers – will be enabled to use automated inference and prediction tools for a broad range of tasks ranging from general-purpose automated maintenance up to AI-assisted custom applications development and deployment.

Artificial Intelligence and more precisely Deep Learning rapid advancement in the past years generate an almost infinite multitude of potential applications with a strong impact in our lives [2]. From business predictive analytics to computer vision, from data security to healthcare it is hard to imagine a field where Artificial Intelligence impact will not be felt in the coming future. One specific area of interest is the area of computer systems and software development and maintenance. In this targeted area, we foresee a number of potential applications some of which are already in research development in various prestigious artificial intelligence laboratories. Among these applications there are two clear directions that we are focused on:

- *Automated legacy application translation using advanced visual inference and automated programming* based on user interface activity. Within this research direction, the main objective is to construct advanced visual recognition systems for artificial scene instances segmentation coupled with classic sequence-to-sequence translation of user actions and visual flow to finally output actual intermediary source code. This intermediary source code must address both the user experience graphical interface and the actual basic functionalities of the user interface control behaviour. An unobvious use case would be the translation of a simple financial management application written for legacy operating systems widely used in '90s and even later to a modern web-based online system that would be uploaded within a cloud computing infrastructure. It is a proven fact that the global business models are gearing towards more collaborative schemes with a clear direction of evolution towards the paradigm of *virtual enterprises* [3]. As such this focus domain is strongly related to the ongoing trend of adopting collaborative intelligence as well as *collaborative decision-making processes and systems*.

- *Intelligent inference of systems maintenance use cases* is the second area where the proposed work's main objective is to advance the state-of-the-art in the area of automated maintenance tools for 3$^{rd}$ party software systems. For this area, we focus on the need to produce intelligent virtual agents capable of replacing the need for remote analysis currently done by software engineering and system administrators. As a use case, we could imagine the maintenance procedure of a client-server system where the end-user interacts with a thin-client user-interface and requires the systems/software engineer's assistance for a potential identified bug. In this case, the proposed end-to-end pipeline could understand the basic behavior and flow of the given user interface (and the potential buggy functionalities) and provide the maintenance team with advanced debugging information.

Both the above-mentioned areas of intervention will be further detailed in the following sections however the main focus of this paper is to present the research findings in the area of artificial (*synthetically generated via computer graphics methods*) scene inference. In this specific area, we started with the most well-known architectures – presented within the *Related Work* section of the paper – we have developed *custom training and validations datasets* and finally, we have identified optimal neural graph architecture for end-to-end artificial scene inference. Aside from the artificial scene inference task we also have included within the scope of work the inference of natural scenes generated by actual hand-drawing of user interfaces mock-ups thus increasing the real and commercial application of the research and experimentation work.

## 2. Related work

The current work relates to the most influential deep convolutional directed acyclic graphs architectures – namely the Inception [4] and ResNet [1] as well as several other architectures such as separable convolution network proposed by Xception [5], and also the fully convolutional model for end-to-end image segmentation FCN [6] based on the straight sequential VGGNet [7] deep neural network.

### 2.1. Inceptions, residual and skip connections

The proposed work is strongly related to de *Inception* architecture developed by Szegedy et al [4] combined with the residual connections proposed by He et al [1] and finally introduced by Szegedy et al in the 4$^{th}$ version of the *Inception* architecture [8]. As it will be presented within the architectural section, we are using a custom version of an *Inception*-residual module interlinked together with modules based on separable depth-wise convolutions fully augmented by residual connections for efficient gradient back-propagation.

*2.2. End-to-end image semantic segmentation*

The proposed neural graph architecture is based on the basic principles of dropping all pooling layers within the convolutional network and inserting larger step convolutions for map width/height reduction and also replacing the dense layers with convolutional. Finally, this results in transforming the entire computational graph into one big fully convolutional directed acyclic graph. As described in the related work by Long et at [6] we replace the final fully connected dense top layers with transposed convolution layers in order to learn upscaling-kernels that will convert the reduced activation volumes from the final convolutions to the initial size and depth of the image. We also use the skip-and-merge strategy in order to fuse lower-level upscaled maps with later, and thus higher-level, upscaled maps.

*2.3. Other similar work in this area*

The field of automatic program generation that strongly relates to the current paper has known many attempts and approaches over the years ranging from systems designed for automatic code generation based on (near) natural language specifications up to source code generation based on an interface mock-up (computer-aided drawing of user-interface mock-up). The closest and newest similar work to our knowledge is the pix2code [9] proposed by Beltramelli T. In relation to this proposed approach we argue that the current work is more generalized as follows:

- in terms of the *target* platform, a cross-platform approach similar to our early work [10]
- in terms of source input the proposed advanced neural model accepts both artificial data (such as screen snapshots) and *hand-drawn natural images* (mock-ups)
- the proposed model generates a dense prediction of the actually observed UI scene (artificial or natural) excluding the need of an RNN-based source code generator and inherited problems such as the potentially erroneous generated code or the need for soft/hard attention (proposed as a future improvement in referenced work). Nevertheless, a second stage decoder can be plugged-in in order to generate the target source code.

## 3. Proposed neural graphs architecture

*3.1. End-to-end trained portable model justification*

One of the direct outcomes of the chosen architecture design is that of being able to deploy the production models on different devices in inference mode the same version of the trained computational graph. The initial challenge consisted of obtaining a neural graph architecture that will be able to forward propagate input scene information with an acceptable performance both on GPU-augmented embedded devices such as Nvidia Jetson family and on mobile devices that do not have GPU parallel computing capabilities (either Android or similar devices).

One of the use cases that we envisioned from the very beginning was a mobile device app that would enable user-interface inference and refactoring. More specifically, the app would enable users with no prior software engineering experience to take a snapshot of a legacy application screen, a website or a simple UI mock-up hand-drawing. Following this initial step, the deployed pre-trained model in the smart-phone device would generate an inference of the snapshot image and produce a UI functional design including multiple visual templates based on inferred functionalities. Finally, the app, using an intermediary module/engine, would generate a simple runtime environment, including minimal views and controllers, for a target web-server and with one or multiple cloud publishing mechanisms would push the functional website where it could be accessed.

*3.2. Neural graph design considerations*

The top-down overview of the *CloudifierNet* can be visualized in *Figure* 1. while the main building blocks of the proposed graph architecture are the custom directed acyclic sub-graph, blocks presented in *Figure 2* and *Figure 3*. The first basic building block based on the *Inception-ResNet* [8] concept is basically a three-column directed acyclic graph with a skip connection that starts with a 1x1 convolution without non-linearity that prepares the input volume and connects it to the final bottleneck final that takes as input the concatenated 3-columns. Within the 3 columns infrastructure, we use simple 3x3 and 5x5 same convolution together with the 1x1 bottleneck convolutions. All convolutions are followed by batch-normalization before the non-linearity. We also employ pre-activation skip connection – a network architecture that does not introduce a non-linearity between any two skip-connections as proposed by He et al in [11].

The purpose of this block architecture is to create a local neural graph topology as proposed by the graph-in-graph architecture of *Inception* coupled with the latest research findings in employing skip-connections. In order to decrease the network size in terms of the number of layers, we use the wide residual network principles described by Zagoruyko et al [12] and increase the volumes of the residual blocks.

The second building block "*DS RES Block*" is based on the concept of separable depth-wise convolution proposed by Chollet [5] in the *Xception* architecture together with the already mentioned skip connections required for the gradient flowing optimization. Each of the *DS RES Blocks* is then upscaled using a transposed convolution 2D (or a so-called *backward convolution*) using a specific fractional stride in order to upscale from respective volume to the initial input volume height and width.

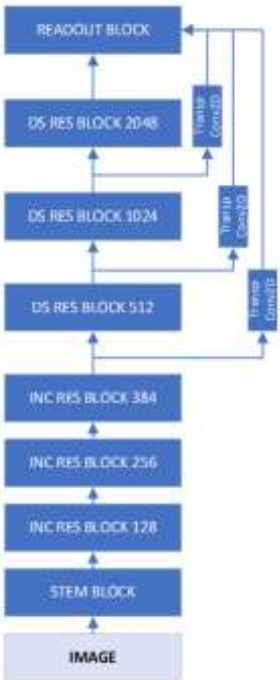
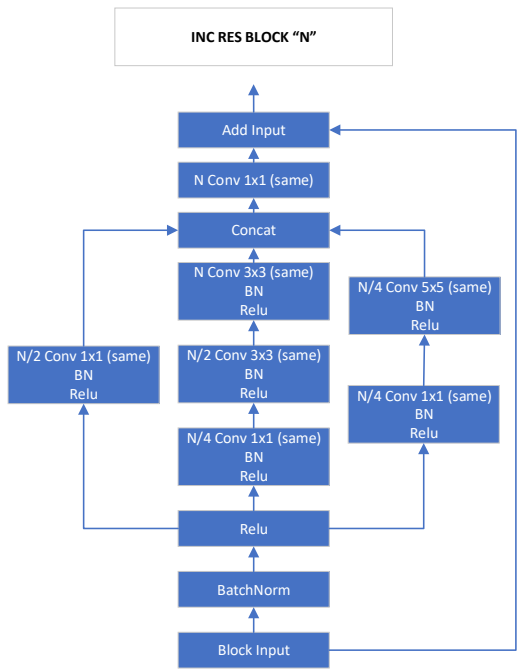
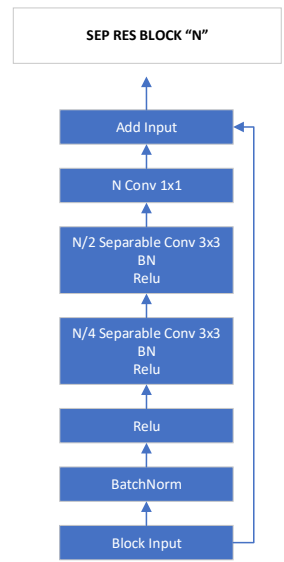

Figure 1– Overall architecture of CloudifierNet

Figure 2– INC RES BLOCK with N output feature maps; last "Add Input" segment represents the residual add operation

*Figure 3– DS RES Block with N output feature maps;*

Within the *Readout-Block* all the upscaled volumes are merged in a *H*W*D* volume where *H* and *W* are respectively the height and width of the input volume and the *D* is the depth of the concatenated volumes from the *DS RES Blocks.* Finally, a dense output map of size *H*W*C* is generated with a simple fully connected layer applied on each of the *H*W fibers* in the *H*W*D* volume, where *C* is the number of inferred classes. This architecture allows the fully convolutional

DAG to be *input volume size agnostic* and to have the capacity to accept any kind of input volume without height or/or width restriction.

As previously mentioned, in order to achieve our proposed goal, we use alternation of the two main building blocks as presented in Figure 1. The architecture has been determined experimentally and has been tested both on natural and artificial images.

The approach of this neural graph architecture allows the obtaining of a good balance between powerful feature detection – based on inception modules with skip connections - and model complexity. Also, a very important goal of the de neural graph design was to maximize the operation parallelism within the computational graph by reducing as much as possible the sequential operations and transferring the complexity to the parallelizable operations.

This is achieved effectively by reducing the length of the computational graph while increasing the so-called depth of each computational node, all this by increasing the number of convolutional kernels in each convolutional operation – either depth-wise separable or classic *Conv2D* operation. As a result, the computational graph can maximize the use of GPU numerical core parallelism both at training time and inference time. For training, we use P5000 family NVidia GPU – as it will be further described within the *Training and Experimentation* section - while at inference time we use both Nvidia Jetson TK1 and the newer TX2 generation.

In terms of complexity, we have an average number of layers compared with various state-of-the-art architectures totaling 109 layers that sum a little over $10^6$ weights for the CloudifierNet model. For initial experiments we used 32bit floats but we plan on further experimenting with reduced size weights (16bit, 8 bit and even extreme 1 bit for inference) based on [13] [14] and with combined float/integer for weights/activations such as the ones proposed by Lai L et al [15].

Finally, the proposed readout layer is a dense prediction *SoftMax* layers where each input pixel is given a probability over all known classes similar to the fully convolutional architecture for image segmentation proposed by Long et al [6]. This structure allows us to further train our models with the negative log-likelihood objective function applied to the entire dense prediction map for all images within a minibatch.

*3.3. Datasets preparation and augmentation*

The proposed neural models have been trained using two different kinds of datasets that we are planning on publishing open-source for further research use:

- the "*artificial*" dataset consisting of software-generated user interfaces controls and actual scenes (full user interfaces)
- the "*natural*" dataset consisting of hand-drawn mockups

The "*artificial*" dataset preparation-augmentation process has been done in multiple iterations. The main aspects that have been considered have been:

a) Operating systems dependent visual aspects of the user interface (OS-theme visual controls). For this issue, we have targeted legacy operating systems and their user-interfaces themes such as Windows 95, 98 and XP.

b) Compiler and development environment-dependent visual control primitives. Narrowing the search to the proposed target operating systems, we have researched several different legacy development environments (such as Borland Delphi 1-3, Visual Basic, FoxPro, etc.) and extracted visual themes and customary user interfaces graphical primitives.

Finally, all "*artificial*" dataset observations have been generated using automated tools that performed the following tasks: (a) visual control generation; (b) automatic image labeling; (c) automatic visual control instance segmentation. The final "artificial" dataset is composed of multiple meta-batches of 3072 observations of 352x352 cropped images with 3 channels, based on the fact that we train our models with various mini-batch size within 32-128 range. As previously mentioned the labels dataset contains both class-per-observation as well as dense pixel-level classes – each image observation has an associated dense pixel-class map and an overall label. We use both coarse labels based on well-known major UI control groups and fine labels where each type of UI visual has multiple sub-classes.

The second dataset, the so-called "natural" dataset consists of hand-drawn examples of user interface primitives or actual user-interfaces mockups. It is important to mention that this dataset is limited in size due to the complicated nature of the hand-drawing and scanning process. Nevertheless, we are using varied image dataset augmentation approaches such as random rotation, random shifting, random channel shifting, random flipping, random rescaling, random cropping in order to dramatically increase the size of our proposed natural image dataset (up to 7 times the original size of the natural dataset). It is important to mention that most of these proposed image data augmentation methods are not required for the artificial dataset due to the actual used image/data acquisition process.

## 4. Training and experiments

### 4.1. Neural models optimization setup

Given the previously presented architecture and datasets with dense-labels per image and considering each image has an *H* x *W* size, for *N* mini-batch images the training objective is to minimize the dense output cross-entropy or more specifically the negative log-likelihood objective function w.r.t. the model parameters *Θ*.

$$\underset{\theta}{argmin} - \frac{1}{N*W*H}\sum_{i=1}^{N}\sum_{x=1,y=1}^{W,H} log\, p_\theta(\hat{y}_{H,W} = y_{H,W}|X,Y) \quad (1)$$

The dense output of the DAG is nothing more than a Softmax function applied to each 1x1xC "fiber" of the final output volume where C is the number of classes.

$$h_{softmax}\left(x_i^{(j)}\middle|\theta, j\in M, i\in N\right) = \frac{e^{\theta^T x_i^{(j)}}}{\sum_k^N e^{\theta^T x_k^{(j)}}} \quad (2)$$

Due to the nature of some of the observations consisting mostly of *background* information and few visual-interface controls we observed a natural tendency of the models to be focused on *well-inferring* background pixels. As a result, we decided to employ in our experiments a modified version of cross-entropy that will enable a down-scaling of the gradients for the well-classified pixels and thus lower the "power" of the background classes. This particular approach has been based on the *focal loss* [16] proposed by Lin et al. We trained the proposed models end-to-end using Adam optimizer on batches of variable size in multiple training experiments. We used an initial learning rate of 0.01 and applied learning rate decay based on monitoring the dev-dataset loss plateauing behavior.

### 4.2. Training infrastructure

As a training environment infrastructure, we used CUDA-based computing infrastructure, however for inference testing we used both GPU and CPU. The training machine CPU capabilities consisted of 32 GB RAM and 2 Xeon processors each with 8 physical cores for total logical core count of 32 cores. Nevertheless, the main training of the proposed models has been done on the machine GPUs summing a total of 4224 CUDA cores provided by a 16 GB Nvidia Pascal 5000 GPU with 2560 CUDA cores and an additional 8GB Nvidia Maxwell 4000 GPU with 1664 CUDA cores.

The benchmarking tests have been performed both on a reduced performance mobile computer using an Nvidia GeForce 940MX with only 2GB RAM and a total of 384 CUDA cores. We have also considered running the inference task and on Nvidia Jetson TX2 device with 256 core Pascal GPU and 8GB of RAM.

*4.3. Optimization process and validation procedures*

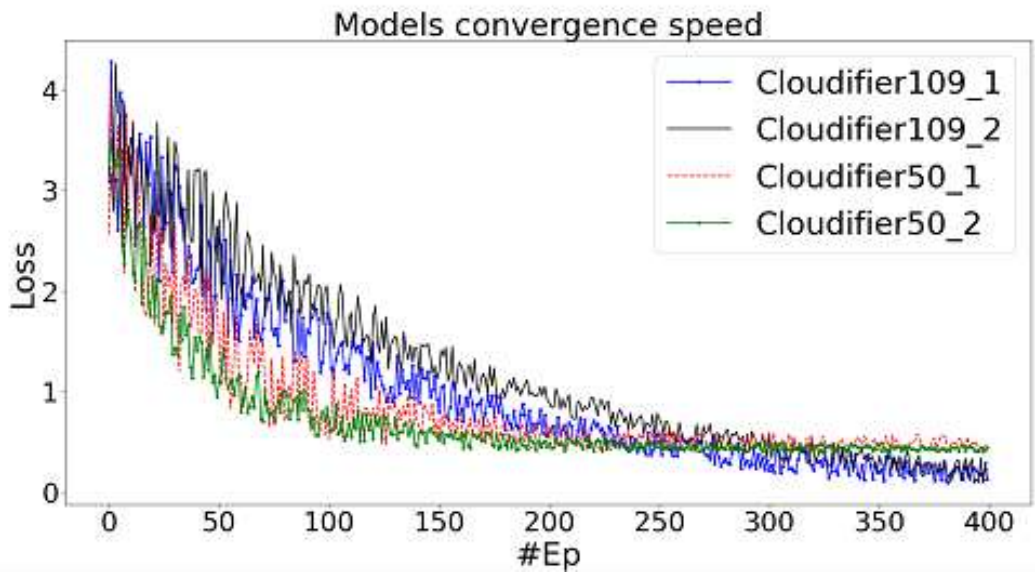

*Figure 4 Presented models convergence speed*

In the proposed pipeline validation experiments we have targeted both types of *visual scenes*: artificial images - actual user interface screen-shots – and hand-drawn user interface mockups. For the experiments, we used several versions of the models that varied both the model required memory/computation capacity and the model training time. We considered this approach in order to cover the potential usage of the model pipeline in smart mobile devices as previously mentioned in the architecture section.

The short graphical summary of the optimization process history is presented in *Figure 4* where we can see the convergence speed of the four proposed test models on the provided combined datasets.

The results of our experiments presented in *Table 1* are based on two different models both following previously described architecture. More specifically, *Cloudifier109* is the 109 layers model presented in the Proposed Architecture section and *Cloudifier50* is a reduced version in terms of modules of the 109 layers version. This second version is basically an almost halved version of the *Cloudifier109* obtained by reducing the total number of modules. For both models, we experimented with a few variations in hyperparameters and thus we have for every two different variants. We also plan to release in the near future the full script source code.

*Table1 – Experimentation results for test dataset (*preliminary results)*

| Model | ArtAcc | ArtRec | NatAcc | NatRec |
|---|---|---|---|---|
| Cloudifier50_1 | *85.1% | *91.2% | *84.3% | *88.0% |
| Cloudifier50_2 | *88.3% | *92.1% | *86.2% | *90.7% |
| Cloudifier109_1 | *95.2% | *97.2% | *93.1% | *95.2% |
| Cloudifier109_2 | *98.4% | *99.7% | *96.1% | *96.1% |

The inference environment for our models has been based on *TensorFlow* [17] and on a specialized inference engine, namely *TensorRT* - a library that facilitates high performance inference on NVIDIA GPUs. Results have been generated by a cross-validation approach and averaged over target environments.

For training/validation/testing split, we decided to retain 7% of the whole joined dataset for validation/testing purposes and 93% has been used for actual model training. Out of the 7% retained dataset, we used 4% for testing and 3% for validation. During the initial training experiments, we additionally used an in-training random validation dataset of 5% of the proposed training data.

The tests and the results have been divided into two categories – the artificial scene inference and the natural hand-drawn mockup scenes inference. In our experimentation results table, *ArtAcc* and *ArtRec* represent the accuracy and the recall for the artificial dataset tests, while *NatAcc* and *NatRec* represent the accuracy and the recall for the natural hand-drawn scenes.

As it can be observed from *Table 1* presented performance indicators, we achieved results that clearly demonstrate the capability of our model to infer user interface functionalities based on static scenes.

## 5. Conclusions

The proposed neural models have achieved beyond state-of-the-art results in the area of user-interface scene inference and a new state-of-the-art status in the area of natural UX mockups inference where we did not identify any clear current state-of-the-art. The results have encouraged us to further propose this work for various production applications within our team, such as stand-alone application migration to collaborative environments or application of this neural graph architecture for user-experience automation (RPA).

Although the proposed model pipeline has proven itself as a powerful approach for the proposed tasks, the end-to-end pipeline is not yet capable of inferring actual high-level functionalities. Thus, one area of further research and improvement of the end-to-end pipeline model is the addition of high-end application functionality inference, albeit only for artificial user-interface video streams. This future proposed work will augment the models with the ability to analyze a video stream presenting an actual user-application interaction and infer actual user experience including high-level process functionality. This will further lead to the facile employment of sequence to sequence models that will generate actual high-level process functionality – rather than the basic one generated by the existing models.

The final and most important conclusion, regarding the ongoing and near-future research based on the current work, is the improvement, extension, and publication of the *CloudifierNet* dataset. We are currently witnessing the wide adoption of RPA solutions, in multiple horizontal and vertical industries, as well as the continuing need for translating legacy systems to new Cloud-based setups. We hope that our current and further research will be able to fuel the field of automated user-interface analysis and application translation, both in academic and industrial environments.